\documentclass[letterpaper]{article} 
\usepackage{flairs}  
\usepackage{times}  
\usepackage{helvet}  
\usepackage{courier}  
\usepackage{url}  
\usepackage{graphicx}  
\usepackage{textgreek}
\begin{document}

\title{Leveraging Evolutionary Algorithms for\\ Feasible Hexapod Locomotion Across Uneven Terrain}
\author{Jack Vice, Gita Sukthankar, and Pamela K. Douglas\\
University of Central Florida\\ Orlando, FL US}

\maketitle
\begin{abstract}
\begin{quote}
Optimizing gait stability for legged robots is a difficult problem. Even on level surfaces, effectively traversing across different textures (e.g., carpet) rests on dynamically tuning parameters in multidimensional space. Inspired by biology, evolutionary algorithms (EA) remain an attractive solution for feasibly implementing robotic locomotion with both energetic economy and rapid parameter convergence. Here, we leveraged this class of algorithms to evolve a stable hexapod gait controller capable of traversing uneven terrain and obstacles. Gait parameters were evolved in a rigid body dynamics simulation on an 8 x 3 meter obstacle course comprised of random step field, linear obstacles and inclined surfaces. Using a fitness function that jointly optimized locomotion velocity and stability, we found that multiple successful gait parameter evolutions yielded specialized functionality for each leg.  Specific gait parameters were identified as critical to developing a rough terrain gait. 
\end{quote}
\end{abstract}

\section{Introduction}
 Small and medium-sized mobile ground robots that operate in urban and rural outdoor environments must cope with terrain that is uneven and cluttered with obstacles. For traversing uneven terrain and scaling obstacles, legged locomotion has several advantages over wheeled and tracked vehicles including enabling omni-directional movement, scaling obstacles higher than the robot's center of gravity, avoiding contact with ground hazards, and fault tolerance.

For biological systems with legged locomotion, evolution by natural selection has favored (in terms of species count) hexapods, which possess the inherent stability of a tripod gait as well as the ability to articulate the front two legs for climbing and manipulation tasks while maintaining a stable stance on the rear four legs.  Effective execution of a hexapod gait generally requires a minimum of 18 degrees of freedom. Thus developing an agile and stable gait for traversing randomly uneven terrain has a large search space that serves as a good candidate problem for evolutionary algorithms, and the application of evolutionary algorithms to the field of gait controller optimization has been a subject of ongoing research ~\cite{lewis1992genetic,kon2020gait}.   

For legged robots on uneven surfaces, a closed loop gait controller that can detect when and with what force the feet contact the ground, and dynamically adjust to the uneven surface, will usually outperform open loop systems~\cite{irawan2012force,koenig2011real}.  However an open loop gait for uneven surfaces is useful for many legged robots which may not have ground contact sensing or motor force feedback. Additionally, some common deformable surfaces, such as vegetation and litter, do not provide ground contact certainty, requiring the use of an open loop gait.  This paper demonstrates a method for evolving an open loop hexapod gait controller capable of traversing uneven terrain and obstacles and presents results on a PhantomX hexapod simulated using the Open Dynamics Engine in Gazebo.

\begin{figure}
    \centering
    \includegraphics[width=0.38\textwidth]{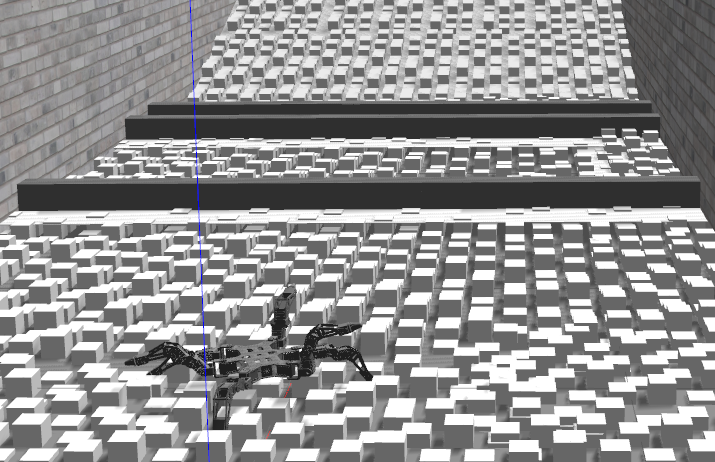}
    \caption{Gazebo simulation environment. Obstructions include a random step field on an incline and linear obstacles.}
    \label{fig:gazebo}
\end{figure}

\section{Related Work}

Evolutionary algorithms have been used to evolve flat surface forward gaits for hexapods in the Open Dynamics Engine; for instance \citeauthor{belter2010biologically} (\citeyear{belter2010biologically}) demonstrated the usage of an evolutionary algorithm in which fitness based selection and crossover was replaced by population density based reproduction. Real value gait parameters were used to encode the genotype. To overcome the reality gap between simulation and the physical robot, they attempted to identify a set of parameters which would induce the most commonality between the simulated gait and the physical robot gait.  However, very little work has been done to evolve a gait controller capable traversing randomly uneven surfaces and scaling obstacles.

An interesting problem in locomotion research is creating gaits for ``injury'' conditions such as complete leg failure ~\cite{manglik2016adaptive} or individual joint failure~\cite{kon2020gait}.
Similar to our work, \citeauthor{manglik2016adaptive} (\citeyear{manglik2016adaptive}) used combinations of sinusoidal waves to govern all leg servo motion. The gait parameters composing the genotype were eight real valued parameters; four leg angles and four distance parameters.  Gaits were evolved using a genetic algorithm with a population of 50 individuals over 40 generations.  Evolution was performed in a simple Matlab mathematical simulation and then gaits were tested on a physical robot. However, when the gaits were transferred to the physical robot, inefficiencies were observed due to the lack of physical dynamics modeling in the simulation (e.g., friction).  This illustrates the problem with conducting the evolution process in low fidelity simulations.  

In an effort to adapt to leg joint failure, \citeauthor{kon2020gait} (\citeyear{kon2020gait}) evolved their 18 DOF hexapod gaits in simulation for 200 generations with fifty individuals per generation, adapting gaits to individual motor loss rather than entire leg loss.  A binary genotype of 540 bits corresponded to fifty-four phenotype decimal gait parameters.  Their method relied on seeding the population with a hand tuned gait; the evolved gaits were able to adapt to single and dual joint failure albeit with many gaits considered spastic.

Rather than evolving a single gait, it may be useful to design multiple gaits for different conditions~\cite{cully2016evolving,li2015parameter}. \citeauthor{cully2016evolving} (\citeyear{cully2016evolving}) 
 evolved a repertoire of hexapod walking gaits by selecting multiple gaits per generation. This was accomplished using a complex fitness mechanism which selected multiple individuals in each population based on the resulting trajectories of the various gaits.  For transfer to the physical robot, gaits developed in simulation using the Open Dynamics Engine were tested on the robot at regular intervals and given a transfer rating. This transfer rating was used to train an SVM classifier.  Once the classifier had been trained with enough transferred gaits, it could predict the likelihood of successful transfer of a given simulated gait to the physical robot.  The methodology produced several hundred functional gait controllers useful for adjusting speed and direction. 


\section{Method}
Simulation environments are often used for evolutionary algorithm work as it is impractical to test thousands of robot control configurations on a physical robot. For our research, the Gazebo simulation environment was used to test evolved gait configurations.  Gazebo can be integrated with the Robot Operating System (ROS) such that Gazebo receives control input from ROS and also provides telemetry back to ROS which can then be published as a ROS topic.  There are a number of physics engines that can be used to power the Gazebo simulation physics.  For this effort, the default Open Dynamics Engine (ODE) was used, though the future work section will discuss additional dynamics engines that could be tested.

The simulated obstacle course shown in Figure~\ref{fig:gazebo} is approximately 3 meters in width and 8.2 meters in length.  The initial 3 meters of the course is composed of a random step field which varies in height from 0 to 7.6 cm followed by a 11.4 x 11.4 cm box beam obstacle and then another meter and a half of random step field followed by a dual box beam obstacle, and finally a 45 degree inclined random step field.  The course is enclosed with walls on either side to prevent the robot from moving off the course and the trial is ended after 90 seconds, or if the robot reaches the top of the incline (8.2 meters).

\begin{figure}
    \centering
    \includegraphics[width=0.25\textwidth]{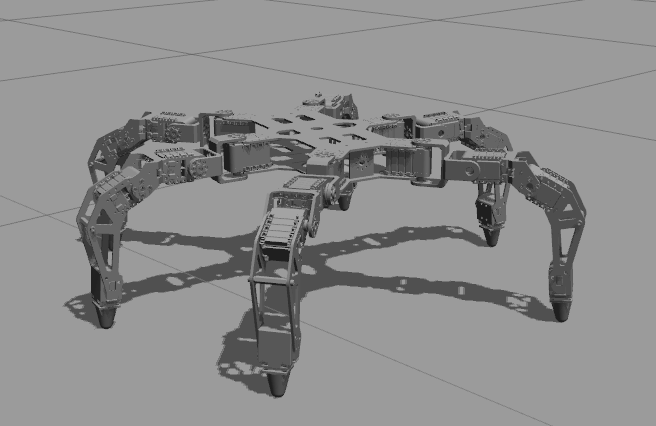}
    \caption{PhantomX hexapod }
    \label{fig:phantom}
\end{figure}

With three degrees of freedom per leg, the PhantomX hexapod (Figure~\ref{fig:phantom}) has a total of 18 degrees of freedom.  The middle leg coxa servos are positioned at a 90 degree angle while the front and rear coxa are at 45 degree angles.  Each leg is 24 cm long fully extended and the thorax is approximately 9 x 20 x 5 cm. The tallest walking gait can achieve a ground clearance of just over 15 cm at the expense of stability and mobility.  Typical stable gait ground clearance is 5 cm. The PhantomX Gazebo model (\url{https://github.com/HumaRobotics/phantomx_gazebo}) was used with an increased mass value of 2.5 kg to more accurately match the mass of the target PhantomX hexapod with a battery and CPU.  For this setup, the phantomx\_walker and Gazebo nodes are created, joint angles are initialized, and then the evolutionary algorithm software creates a controller node. This node sends joint angles to the PhantomX controller which in turn communicates the model movements to the simulation. Lastly, a listener node receives ground truth of the robot's position and orientation to be evaluated by the fitness function.  

For the hexapod control, the coxal servos are set to a constant tripod joint angle trajectory using a continuous sinusoidal wave with the ROS clock as the input value.  A hexapod tripod gait is a symmetric gait in which the front and rear legs on a given side will lift and move forward while the middle leg pushes rearward on the ground. Then the front and rear legs touch down and push back while the middle leg lifts.  Both sides are timed so that there are always three feet on the ground, one on one side and two on the other.  The front and rear coxal servos move forward in unison while the middle servo moves rearward, with the opposite trajectory on the opposing side.  Coxal servo range values were set to maximize range of motion without collision and limited to reaching directly forward and directly rearward. The evolution of femur and tibia parameters is such that the two joint angles follow a sinusoidal wave with either the same period or double the period of the coxal trajectory.  Assuming a symmetric gait only requires that half the total gait parameters need to be evolved as one side is simply a phase shift from the other side for all joint angles.   Consequently, there are 8 parameters for each leg for a total of 24 real value gait parameters composing the genome.  For each leg, the parameters are as follows: femur period, femur phase, femur range, femur vertical shift, tibia period, tibia phase, tibia range, and tibia vertical shift.  The period value was either 1 or 2, phase value ranged from 0 to 2, servo range from 0 to 1.7 and vertical shift valued -1 to 1.

Similar to the joint angle trajectory approaches~ \cite{manglik2016adaptive}, the trajectory of each leg joint is defined by a periodic function \textgamma of the amplitude \textalpha, phase shift $\phi$ , period multiplier \textit{p}, time \textit{t}, and the vertical shift \textit{v}. 
\begin{center}
\( \gamma ( \alpha , \textit{p}, \textit{t}, \phi , \textit{v} ) =  \alpha \cdot \cos(( \textit{p} \cdot \textit{t} ) + \phi ) + \textit{v} \)
\end{center}

This sinusoidal wave function for gait control was designed to achieve a number of goals. With the intent to transfer the gait controller to the physical robot, and because the physical dynamics accuracy of the PhantomX Gazebo model is not known with certainty, a complex wave function ~\cite{li2015parameter}, could evolve into a gait in simulation that would be beyond the maximum angular velocities of the physical PhantomX servos, causing the servo trajectory to lag behind that of desired.  This effect could be compounded given the unknown update rates for the main controller loop when running on the robot’s CPU.  Because ROS is a distributed architecture, the simulator is running on a separate CPU from the controller, allowing the controller loop to execute quickly.  It must be taken into account that the controller loop may not update as quickly while running on the robot's ARM processor which will also be processing sensor data.

The performance of EAs rest on fitness functions which evaluate and select individuals, z, in combination based on their evolutionary utility for future generations. In robotic locomotion tasks, distance successfully traveled, orientational stability, and speed are crucial elements of performance. Here, we combined these performance metrics into a fitness function, designed to jointly maximize velocity, $v$, and negative orientational displacement, $D$, to quantify locomotion stability:

\[
fitness=\max_{z_j \ \in {C}}f(v,D_{p})
\textrm{where} \quad D_{p}=\sum_j-\frac{|\partial{D_j}|}{\partial{D_{j_0}}}
\]

Here, $D_p$ is the negative orientational displacement pooled across dimensions with respect to initial orientation conditions. Note, this could also be cast as a maximization, minimization problem.

There are two basic methods to initialize a population for the first generation of an evolutionary algorithm.  The first is generating random (Gaussian or uniform) legal values for each parameter. The second is to start with a seed individual and randomly mutate the individual to form the first population.  For this effort, the initial population was a Gaussian mutation of a hand tuned all terrain gait.  The seed individual itself was not included in the first generation. For each new generation, the fitness function was used to evaluate all individuals and then two parents were chosen using tournament style selection. Specifically, the first parent was selected as the highest fitness individual from the first half of the population and the second parent was the highest fitness individual from the second half of the population.  Once parents were selected, a new population is produced in the following manner. Each new individual has a 0.5 probability to be either generated by Gaussian mutation of the higher fitness parent or generated by a two point random genetic crossover of both parents.  For those mutated from a single parent, the mutation probability was 0.4 for each parameter.  Elitism was not employed, so parents did not survive to subsequent generations.  The two point crossover function used for the evolutionary algorithm was imported from the DEAP evolutionary computation framework (\url{http://github.com/deap/deap}). 

As to be expected with random mutation, many individual gaits were found to be extremely unstable during initial testing.  To manage this, maximum orientation thresholds were imposed such that an individual gait trial would terminate early if the gait proved excessively unstable or reversed in direction.   During the trial for each individual,  the published (ROS topic) quaternion values were used to determine drastic changes in orientation and thus end the trial early if the individual went outside specific orientation parameters.  Maximum orientation parameters were set at 45 degrees roll, 70 degrees pitch and 90 degrees yaw.  Additionally, if the gait drove the robot in reverse for more than 0.2 meters, the trial was ended.  The average rate of change in the \textit{w} quaternion value was used to measure orientation stability of the gait. This stability accounted for 5\% of the overall fitness score.  The time period for each trial was set at 90 seconds.  Lateral displacement was not explicitly penalized, as a real world gait controller would provide steering capability and any open loop gait is likely to veer of course a bit in a random step field.  The overall linear directional fitness provided implicit penalty to lateral displacement, and though it may possible for an erratic gait to achieve higher linear directional fitness by bouncing off of the walls, the orientation stability component of the fitness value would explicitly penalize such a gait. We executed the evolutionary algorithm with a population of 200 individuals for 23 generations.  



\section{Evaluation}

For comparison, the hand tuned, fast, flat surface gait was able to traverse the 8.6 meters on flat ground in 34.8 seconds.  When traversing the random step field, this initial tripod gait only achieved 1.28 meters in the same time period.  The evolutionary algorithm evolved a gait controller which was able to traverse over 6 meters of random step field and obstacles over 90 seconds.

\begin{figure}[h]
    \centering
    \includegraphics[width=0.45\textwidth]{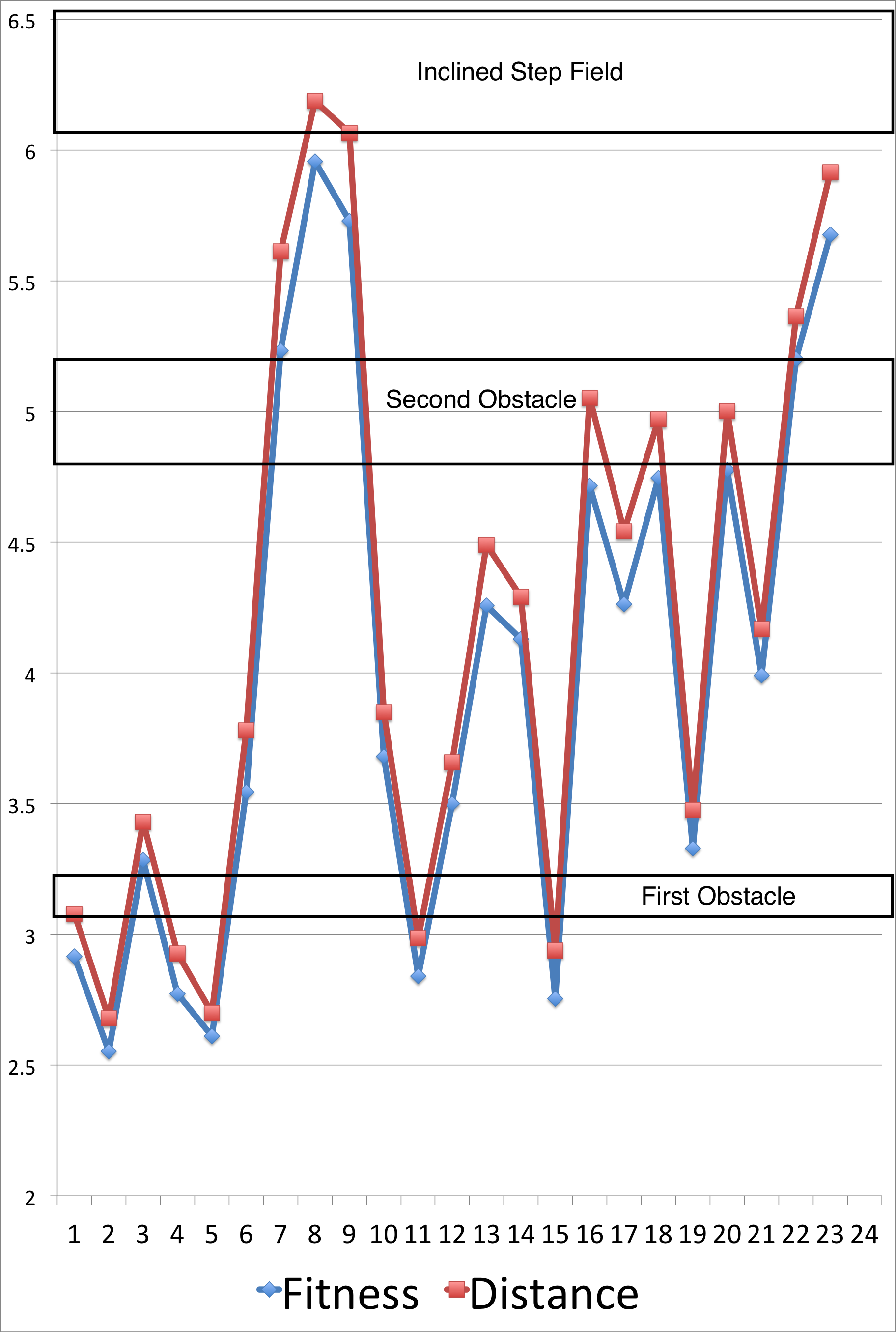}
    \caption{Fitness (blue) and distance traveled (red) by generation (x-axis). The highest distance achieved in generation 8 had relatively low stability.}
    \label{fig:Fit-Dis}
\end{figure}

Fitness results indicate that the evolved gaits were able to reach the first obstacle starting with the 1st generation and then didn\textquotesingle t find a good obstacle climbing gait until the 6th generation (Fig~\ref{fig:Fit-Dis}). From the 5th generation to the 8th, the algorithm achieved a rapid improvement in fitness with the fitness scoring diverging slightly from the distance achieved once the hexapod passed the 5 meter mark, likely due to the obstacles at that point.  Distance achieved for generation 8 was 6.2 meters which is just starting to climb the step field incline.  Fitness then sharply dropped off for the next few generations; the robot was unable to traverse the second obstacle until generation 22 and 23, with generation 23 reaching the second furthest distance of 5.9 meters.  If elitism had been employed, in which parents survive to the next generation, the generation 23 gait solution would likely not have been discovered, limiting the overall search space.

\begin{table}
  \begin{tabular}{ | l | c | c | c | c |}
    \hline
     & Gen. 8 & Gen. 23& Diff. & \% Range\\ \hline
    Fitness & 6 & 5.7 &.3 &5\%\\ \hline
    Distance &6.2 &5.9 &.3 &5\%\\ \hline
    Front Femur Phase&6.28&6.28&0.00&0\%\\ \hline
    Front Femur Range&0.41&1.27&0.86&51\%\\ \hline
    Front Femur Shift&-0.27&-1.00&.73&37\%\\ \hline
    Front Tibia Phase&2.29&2.89&0.60&10\%\\ \hline
    Front Tibia Range&1.50&1.36&0.14&8\%\\ \hline
    Front Tibia Shift&0.71&0.81&0.11&5\%\\ \hline
    Middle Femur Phase&0.46&6.28&5.82&7\%\\ \hline
    Middle Femur Range&0.97&1.08&0.11&6\%\\ \hline
    Middle Femur Shift&-0.67&0.31&0.97&49\%\\ \hline
    Middle Tibia Phase&0.94&1.36&0.41&7\%\\ \hline
    Middle Tibia Range&0.05&0.73&0.67&40\%\\ \hline
    Middle Tibia Shift&0.41&0.54&0.13&7\%\\ \hline
    Rear Femur Phase&6.23&.71&0.52&10\%\\ \hline
    Rear Femur Range&0.12&0.62&0.50&29\%\\ \hline
    Rear Femur Shift&-0.75&-0.87&0.13&6\%\\ \hline
    Rear Tibia Phase&1.62&6.03&4.41&70\%\\ \hline
    Rear Tibia Range&0.14&0.87&0.73&43\%\\ \hline
    Rear Tibia Shift&0.00&-0.21&0.21&11\%\\ \hline
  \end{tabular}
  \caption{The two best gaits compared.  The scalar value difference is shown along with the percentage difference.  The large difference in gait values likely indicates the generation 23 gait was toward a different optima in search space.}
\label{table:best}
\end{table}

Table 1 shows the gait parameters for generation 8 and generation 23, the highest fitness individuals.  The period multiplier value is not shown as it converged to one in every generation but the first.  The average difference between the two gaits is 22\% with 9 of the 18 parameters values within 10\% of each other.  The front tibia kinematics appear to be most similar between the two gaits at only 20\% difference, while the rear tibias have the most difference at 41\%.  With only a 5\% difference in fitness and distance between the two gaits, there are no obvious indications that gait 23 was converging toward the same optima as gait 8. 

\begin{figure}[h]
    \centering
    \includegraphics[width=0.38\textwidth]{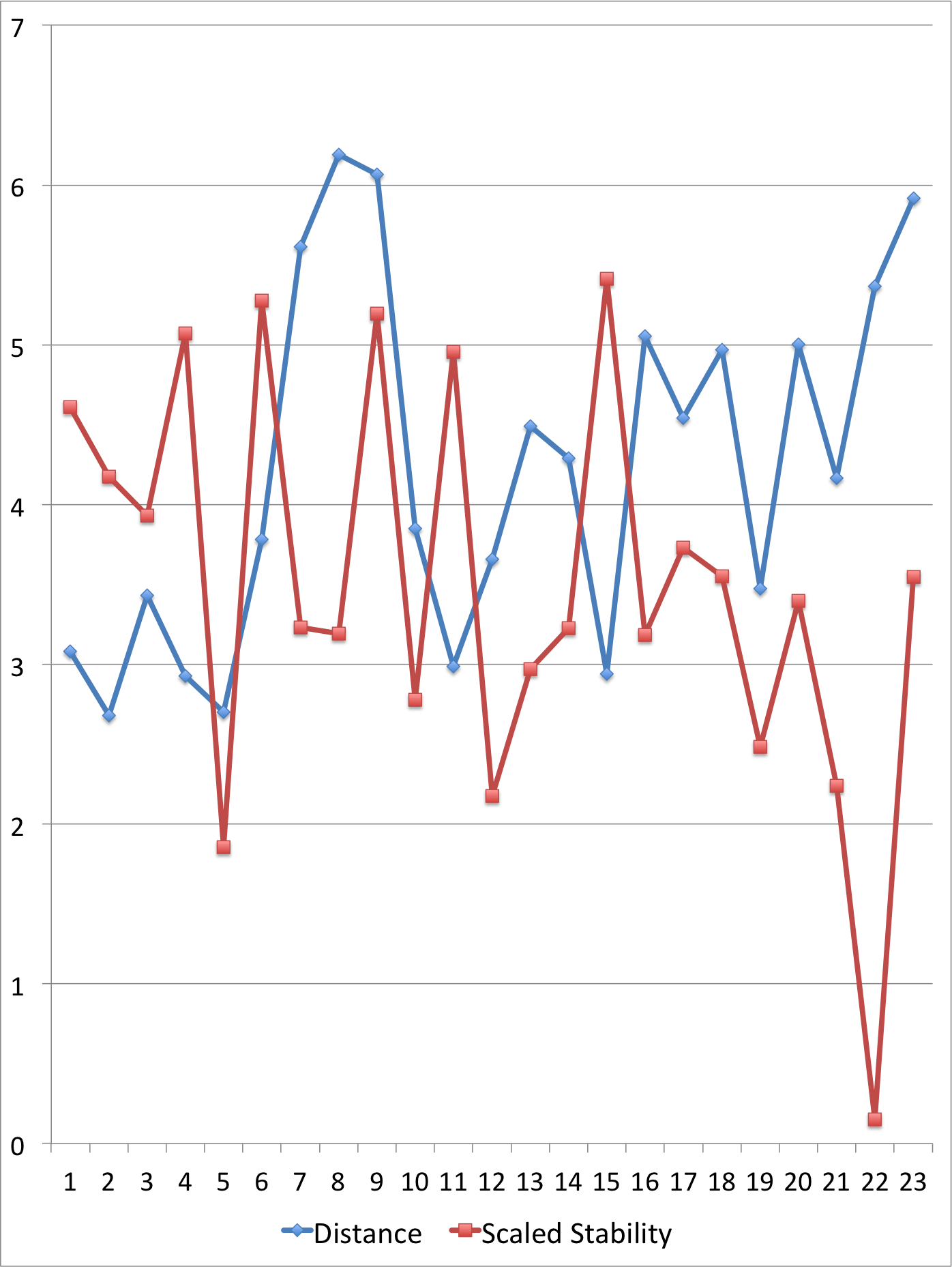}
    \caption{Distance achieved in meters (blue) and scaled orientational stability value (red) is shown for each generation (x-axis).}
    \label{fig:Stab-Fit}
\end{figure}

Figure~\ref{fig:Stab-Fit} shows distance achieved versus scaled stability.  A higher stability value in the chart indicates higher orientational stability.  The stability (average change in quaternion \textit{w}) over the generations indicates that the initial high fitness of generation 8 was not as stable as the solution found in generation 23, further evidence of the different location in search space between the two solutions.  Traversing the box beam obstacles has the highest adverse impact on stability, in that the hexapod encounters a vertical obstacle which causes rapid changes in pitch when traversing.  Three of the five highest stability gaits were the ones that stopped at or just before the first obstacle (gaits 4, 11 and 15).  Consequently, the main stability challenge was the dual box beam obstacle which was configured with the first beam starting at 4.75 meters and the next beam at 5.2 meters, leaving only a 35cm gap between box beams. The PhantomX hexapod neutral stance has a minimum of 40cm between front and rear feet so the dual box beam forms a single obstacle.  For gaits 16, 18 and 20, encountering this obstacle likely adversely affected their stability score as well as halting any further directional progress.

\begin{figure}[h]
    \centering
    \includegraphics[width=0.4\textwidth]{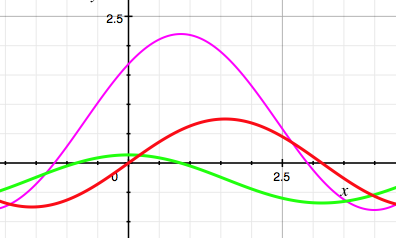}
    \caption{Front leg angular trajectories.  Red is the coxa, green is the femur and blue is the tibia, with higher coxa values being servo forward and higher values for the femur and tibia both being servo up.}
    \label{fig:front}
\end{figure}

Figure~\ref{fig:front} shows the right front leg angular trajectories for the highest fitness gait.
 As illustrated, it is evident that just before the coxa drives rearward, the femur starts its downward stroke, followed quickly by the tibia. Next, the femur and tibia start their upward stroke before the coxa reaches its rearmost point.  The tibia range (amplitude) of 1.5 is much greater than that of the femur range 0.4 which is in stark contrast to the hand tuned all-terrain seed gait which had a tibia range of only 0.4 and a femur range of 0.8.  Furthermore, the tibia wave is shifted higher, giving the tibia the ability to reach up and over obstacles.



Unlike the front legs, where the tibia had a much higher range of motion than the femur, in the middle legs the femur is doing most of the work while the tibia is angled up slightly but with very little range of motion.  For the rear legs, femur and tibia have evolved a very low range of motion.  With the femur angled down (horizontal shift) and the tibia in phase with the coxal, it would appear the rear legs are acting more to stabilize the hexapod orientation while still providing ground clearance versus any significant contribution to  forward momentum.  




\begin{figure}
    \centering
    \includegraphics[width=0.45\textwidth]{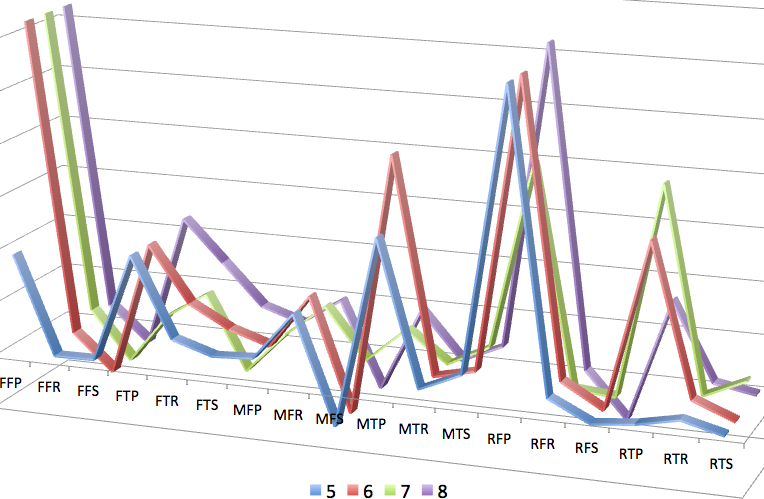}
    \caption{Change in gait parameters (generation 5 through 8). The nearest data is generation 5 with Front Femur Phase (FFR) on the left, then Front Femur Range (FFR), then Front Femur Shift (FFS), Front Tibia Phase (FTP) and so on for middle and rear values. }
    \label{fig:FiveEight}
\end{figure}

Figure~\ref{fig:FiveEight} shows the parameter data from generation 5 through 8; it reveals a rapid increase in fitness, going from the second lowest fitness to the highest in only 4 generations. The largest increase in fitness was observed between generations 6 and 7 where we see a substantial decrease in middle tibia phase which may be correlated with the large change in fitness.


\begin{figure}[h]
    \centering
    \includegraphics[width=0.45\textwidth]{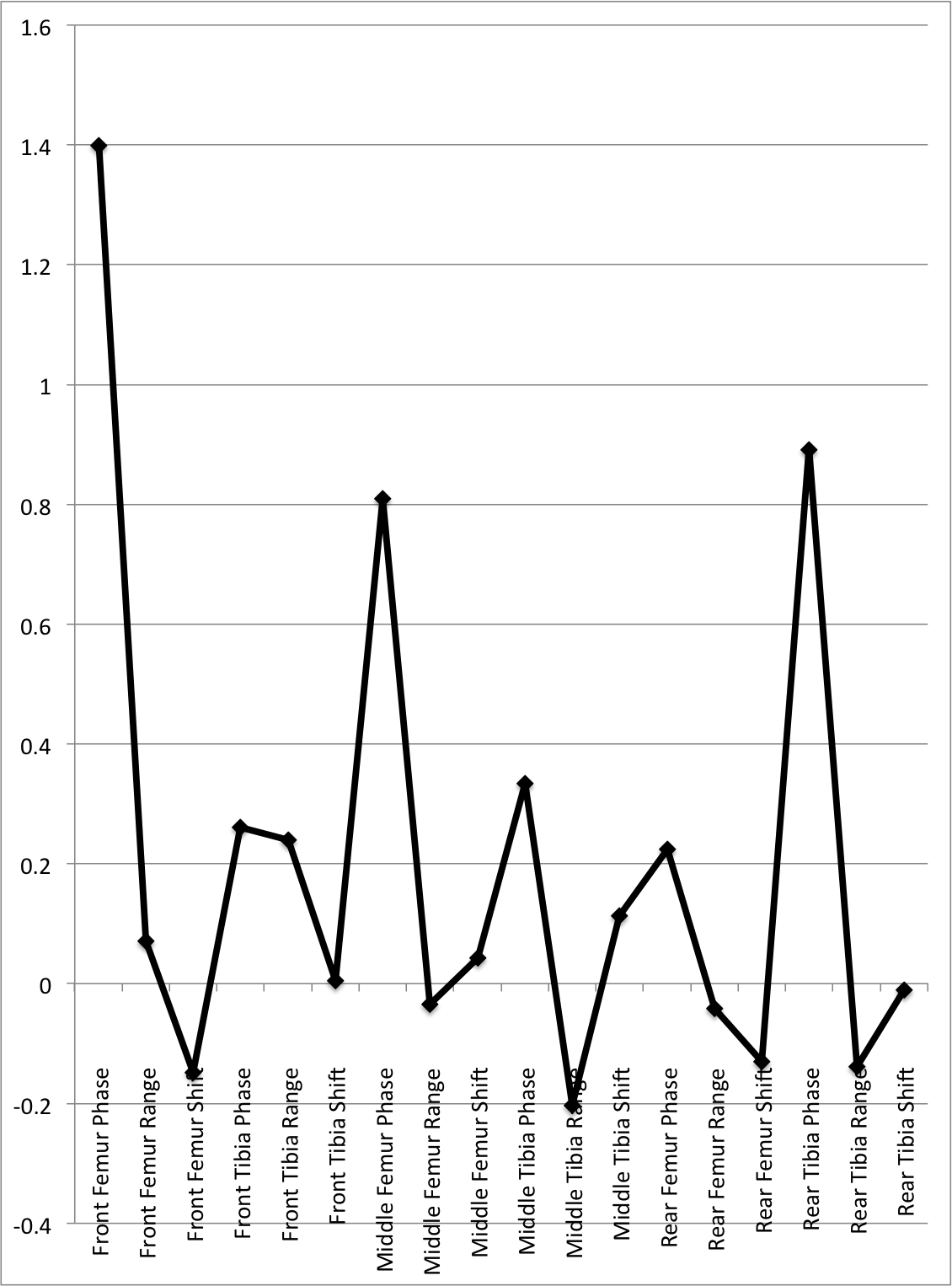}
    \caption{Gait covariance with distance.  This analysis reveals that the front femur phase, middle femur phase, and rear tibia phase parameters are highly correlated with distance achieved.}
    \label{fig:covariance}
\end{figure}

By examining the covariance of each parameter over all generations with respect to distance traveled (Fig.~\ref{fig:covariance}), it is evident that front femur phase, middle femur phase and rear tibia phase are key parameters for a successful adverse terrain gait evolution.  Front femur vertical shift, middle tibia range, rear femur vertical shift and rear tibia range are the least correlated with distance traveled.  Since the phase shift parameters control the relationship between a given joint movement cycle with respect to the coxa, having this timing off would greatly affect vehicle movement (e.g., if the coxal joint is moving rearward and the femur has the leg in the air versus on the ground). 


\section{Conclusion and Future Work}
Using an evolutionary algorithm, an open loop gait controller was evolved to enable a hexapod robot to traverse uneven terrain and obstacles in simulation.  Using a population of 200 individuals over 23 generations, the evolutionary algorithm evolved a gait that was able to traverse just over 6 meters of random step field and linear obstacles in 90 seconds.  The highest fitness score was actually reached in the 8th generation, then fitness lowered for the population and eventually climbed back up to 96\% of the earlier solution. 

For the given hexapod simulation model and physics engine, evolving a multi-terrain gait for a high degree of freedom legged robotic platform was successful in fewer generations than anticipated.  A gait was evolved that could traverse random step fields, box beam obstacles and a inclined step field while also maintaining a degree of orientational stability. Future research into characterizing highly transferable multi-terrain gaits could allow a multitude of obstacle traversing gaits to be developed in simulation for successful transfer to the physical robot. Furthermore, considering the correlation of changes in gait parameters with respect to  performance should inform additional refinement towards the most efficient strategy for evolving robotic gaits.
 
Evolving the gait in simulation is only useful if it is transferable to an actual hardware platform.  Given this, the next step is to test the various peak performing gaits on the actual hardware of the PhantomX hexapod.  From the fitness graph, it is evident that there may be more than one capable gait worth transfer testing.
 
Selecting and fine tuning the various evolutionary algorithm techniques and choice of parameters has a significant impact on algorithm performance.  The practice of meta-evolution, in which these variables are themselves evolved and optimized, enables much more efficient genotype-to-phenotype mapping as shown by previous work ~\cite{scott2015learning}.  Specifically, a linear pleiotropic encoding would enable a single genotype value to alter multiple gait parameters and discover solutions more efficiently.

A future study could also implement different rigid body dynamics engines in Gazebo such as Bullet, DART or Symbody, and then test transfer to the physical platform to see which dynamics engines are evolving the most transferable gaits.  Additionally, implementing machine learning to classify gaits based on the the quality of each gait transfer from hardware to simulation  ~\cite{cully2016evolving} could also help overcome the ``reality gap''.   

Anther consideration for future research is a repertoire of gaits approach in which a gait controller is evolved separately for each obstacle class.  These gaits could then be bred as seed genomes for an initial population, in the interest of discovering new multi-obstacle hybrid gaits.

Finally, Gazebo supports a contact sensor module, which reports collisions and forces between objects.  This module could provide the means to evolve a closed loop gait controller if the contact sensor were added to the tibia/foot.  A dynamic controller that could maximize agility as well as minimize contact forces would help to increase power efficiency, which is a significant challenge for legged robotics, especially over rough terrain.

\bibliography{biblio}
\bibliographystyle{flairs}
\end{document}